\documentclass[pdflatex,sn-mathphys,iicol]{arxiv}

\usepackage{
graphicx,
amsfonts,
textcomp,
program,
amssymb,
listings,
rotating,
booktabs,
algorithmicx,
mathrsfs,
url,
appendix,
etoolbox,
algpseudocode,
multirow,
algorithm,
amsmath,
hyperref,
manyfoot,
hhline,
pifont
}
\usepackage{subcaption}

\usepackage{booktabs}
\usepackage[table]{xcolor}
\usepackage{balance}

\usepackage{array}
\usepackage[export]{adjustbox}

\usepackage{multirow}
\usepackage{algpseudocode}
\usepackage{setspace}
\usepackage{bm}

\usepackage{colortbl}
\definecolor{color3}{gray}{0.95}

\jyear{2023}
\theoremstyle{thmstyleone}

\theoremstyle{thmstyletwo}

\theoremstyle{thmstylethree}

\begin{document}

\title[Article Title]{TextWand: A Unified Framework for Scene Text Editing}

\author{\fnm{Shuyu} \sur{Wang}}

\author{\fnm{Zhile} \sur{Guan}}

\author{\fnm{Hongxiu} \sur{Chen}}

\author{\fnm{Yule} \sur{Duan}}

\author{\fnm{Weiqi} \sur{Li}}

\author{\fnm{Xin} \sur{Shan}}

\author{\fnm{Ronggang} \sur{Wang}}

\author{\fnm{Jian} \sur{Zhang}\textsuperscript{*}}

\affil{\orgdiv{School of Electronic and
Computer Engineering, Peking University}, 
\orgaddress{\city{Shenzhen}, \country{China}}}
    \abstract{We propose \textbf{TextWand}, a general-purpose framework that unifies scene text removal, generation, and replacement into a single model. By decomposing complex editing tasks into the atomic primitives of rendering and erasure, TextWand achieves precise control over both text appearance and background integrity. Specifically, we introduce a novel design, Overlay-Reference Positional Encoding (ORPE), to enforce pixel-level layout fidelity and exemplar-driven style control, alongside a new strategy, Region-Adaptive Suppression (RAS), to ensure clean text erasure. To address the absence of a comprehensive benchmark for general-purpose scene text editing among existing single-task datasets, we construct TextWand-Bench. Extensive experiments demonstrate that TextWand outperforms existing leading open-source and closed-source models by delivering superior text content accuracy, layout and style consistency, and overall image quality across scene text removal, generation and replacement tasks. }


\maketitle
\begin{figure*}[t]
  \centering
  \vspace{-18pt}
  \includegraphics[width=\linewidth]{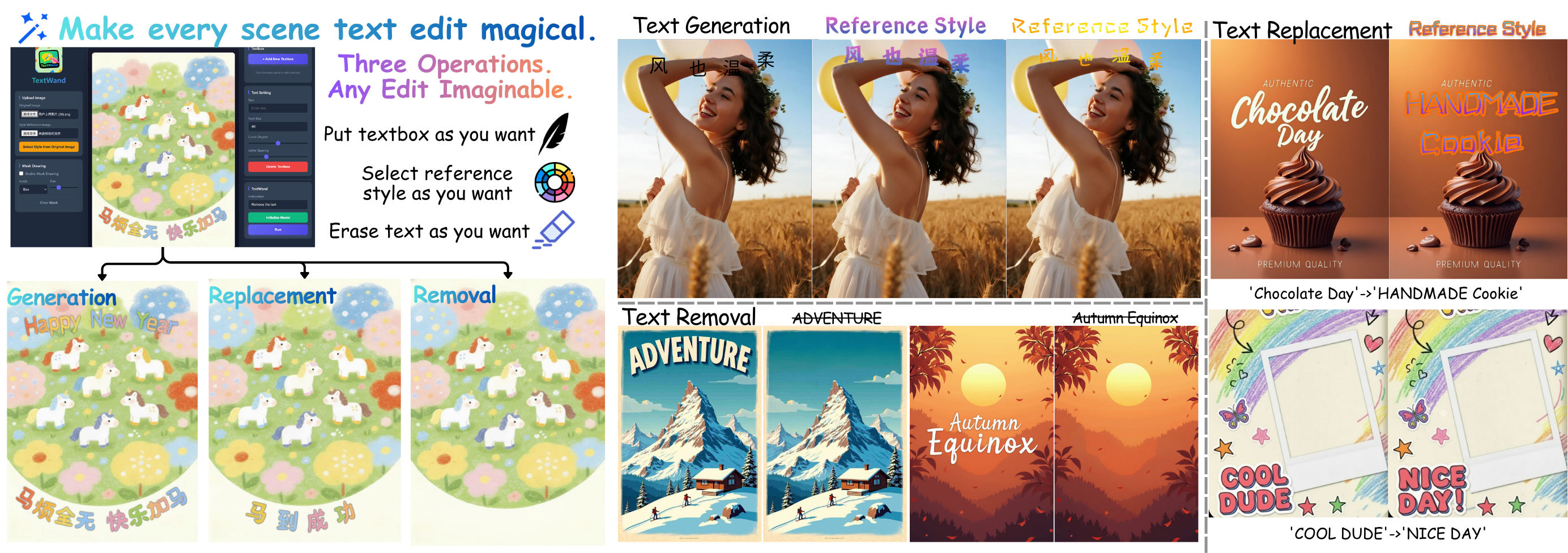} 
  \vspace{-15pt}
  \caption{\textbf{Unified and Interactive Scene Text Editing with TextWand.} 
\textcolor{blue}{Left}: Our interactive interface allowing intuitive control via textboxes, style references, or erasure masks. 
\textcolor{blue}{Right}: TextWand demonstrates exceptional versatility and highly realistic results across three core tasks: scene text generation, replacement, and removal, achieving precise control over layout and style in diverse image domains.}
  \label{fig:teaser}
\end{figure*}
\section{Introduction}\label{sec:intro}
Scene text editing aims to modify textual content within real-world images or rendered graphics (e.g., photos, posters, and screenshots) while seamlessly preserving the original visual context~\cite{shu2025visualintroduction,yang2023selfintro,lee2022surprisinglyintro,radford2021learningintro}. This task is central to high-frequency design workflows such as e-commerce creatives, advertising materials, posters and banners~\cite{bai2024intelligentintroduction}, where the same visual asset must be updated repeatedly across languages, seasons, prices, and promotions. Specifically, it encompasses three fundamental operations: scene text removal, generation, and replacement.  

Traditionally, these tasks have relied on layer-based design software (e.g., Adobe Photoshop) and tedious manual labor. 
Such workflows are effective when the original editable source file is available, where text remains a separate vector layer and the corresponding font files, layer styles, and background content can be directly accessed. However, many visual assets encountered in real-world scenarios are flattened raster images, such as product photos, scanned posters, screenshots, online banners, and photographs of physical signs. In these cases, the text is no longer separable from the image: the original font may be unknown or unavailable, the text layer cannot be extracted, and the glyphs are often entangled with shadows, highlights, perspective distortions, material textures, or complex background patterns. Consequently, a conventional  pipeline is often insufficient. 
While recent generative models~\cite{wu2025qwenimage,labs2025flux1kontextflowmatching,esser2024scaling,cao2025hunyuanimage} have significantly advanced general image editing, they struggle with the unique complexity of scene text. Unlike typical visual objects, text requires strict glyph accuracy and style fidelity. Specifically, a successful edit must simultaneously satisfy multiple stringent constraints: (1) cleanly erasing legacy text without artifacts; (2) rendering new text legibly; (3) adhering to complex layouts (e.g., slanted or curved); (4) matching original typography (e.g., color, shadow, texture); and (5) preserving the background seamlessly. Existing generative editing models still fall short in jointly fulfilling these requirements.

Although recent methods~\cite{fonts_iccv25,fluxtext_2505,anytext2_2411,11095155GlyphMastero} attempt to tackle scene text editing, they struggle to fully satisfy these stringent constraints. 
Most prior works are highly specialized. They typically focus exclusively on scene text generation~\cite{fluxtext_2505,tuo2023anytext,anytext2_2411,zhao2025utdesign,zeng2024textctrl} or scene text removal~\cite{removal1,removal2,9609970remove,10214243remove}, or they concentrate solely on style transfer for existing text~\cite{fonts_iccv25}. Specifically, the limitations of current generative approaches in handling these constraints manifest in three primary aspects. \textbf{Firstly}, existing approaches~\cite{zhao2025utdesign,easytext_2505} cannot seamlessly integrate scene text removal, generation and replacement within a single coherent framework.
\textbf{Secondly}, most of them~\cite{fonts_iccv25,anytext2_2411,fluxtext_2505,omnitext_2510} struggle to achieve joint control over text layout and typographic style during scene text generation and replacement. Most existing methods primarily focus on matching typographic styles but fall short in managing the spatial typesetting of the text within the image. Furthermore, even the approaches that attempt layout control~\cite{liu2024glyphlayout,chen2023textdiffuser,chen2024textdiffuser2} often fail to achieve precise, character-level spatial alignment. Consequently, it remains highly challenging for current models to impose an arbitrary reference style while strictly adhering to a user-specified layout.
\textbf{Thirdly}, the erasure of original text and the rendering of new text are frequently entangled.
Scene text replacement requires the model to simultaneously suppress the visual evidence of the old text and accurately generate new glyphs. Without an explicit mechanism to separate the erasure process from the rendering process, these two actions severely interfere with each other. As a result, models often leave visible ghosting artifacts from the original text or inadvertently damage the surrounding background textures when forced to overwrite the region, making it imperative to decouple these conflicting operations into distinct primitives.

In this paper, we propose \textbf{TextWand}, a versatile framework that unifies text removal, generation, and replacement within a single model, while enabling reliable joint layout and style control in complex real-world images. We achieve this by decomposing the complex editing process into two atomic primitives: rendering and erasure. To concretely support these primitives, we present \textbf{O}verlay-\textbf{R}eference \textbf{P}ositional \textbf{E}ncoding (ORPE) to enforce tracing-level layout fidelity and exemplar-driven styling for the rendering process, and \textbf{R}egion-\textbf{A}daptive \textbf{S}uppression (RAS) to instantiate the erasure primitive ($\mathcal{P}_{erase}$), effectively suppressing residual text evidence to ensure clean background reconstruction across diverse editing tasks. Additionally, we introduce a progressive curriculum training strategy to ensure stable training. We further design an intuitive user interface to operationalize these primitives, allowing users to explicitly specify their erasure intent via a brush mask and their rendering intent via a textbox with a style exemplar. Finally, to address the absence of standardized evaluation for this unified task, we construct TextWand-Bench, a comprehensive benchmark comprising 1500 high-quality test cases meticulously categorized into three core tasks: scene text removal, generation, and replacement. Extensive experiments demonstrate that TextWand achieves state-of-the-art performance across all three tasks, significantly outperforming existing baselines in both text content accuracy, layout and style consistency, and overall image quality.

In summary, our contributions are as follows:
\begin{itemize}
\item We propose \textbf{TextWand}, a general-purpose framework that unifies text removal, generation, and replacement. By formulating editing as rendering and erasure primitives, it enables precise control across diverse scenarios.
\item We design two complementary mechanisms: \textbf{O}verlay-\textbf{R}eference \textbf{P}ositional \textbf{E}ncoding (ORPE) for joint layout-style rendering, and \textbf{R}egion-\textbf{A}daptive \textbf{S}uppression (RAS) for erasure, supported by a progressive training strategy.
\item We construct the first comprehensive benchmark dedicated to unified scene text editing, systematically evaluating text removal, generation, and replacement under strict layout and typography constraints.
\item We develop a user-friendly interaction interface based on brush and textbox controls that makes editing intent explicit and reproducible. 
\item Extensive experiments demonstrate the effectiveness of TextWand and validate the role of each proposed component.
\end{itemize}
\textbf{Organization of This Paper.} 
The remainder of this paper is organized as follows. 
Sec.~\ref{sec:related} reviews recent progress in controllable image editing and scene text editing, with a particular focus on the limitations of existing methods in unified text removal, generation, and replacement. Sec.~\ref{sec:method} introduces the proposed TextWand framework, including the unified formulation based on rendering and erasure primitives, the Overlay-Reference Positional Encoding (ORPE), the Region-Adaptive Suppression (RAS), and the progressive curriculum training strategy. Sec.~\ref{sec:experiments} presents the construction of TextWand-72K and TextWand-Bench, describes the experimental settings, and reports quantitative, qualitative, user study, and ablation results across scene text removal, generation, and replacement tasks. Finally, Sec.~\ref{sec:conclusion} concludes the paper and discusses future directions.


\section{Related Work}
\label{sec:related}
\subsection{Controllable Image Editing}
Diffusion-based generative models~\cite{rombach2021highresolutioninpainting,flux2024,podell2023sdxlgenerative,esser2024scaling,team2025zimage,wu2025qwenimage,seedream2025seedreamedit} have substantially advanced image editing by enabling high-fidelity modifications while preserving realism and global consistency. To achieve this, recent works havde explored various control mechanisms, ranging from natural language instructions~\cite{liu2025step1x-edit_textinstruction,wang2025editinfinity_textinstruction,brooks2023instructpix2pix__textinstruction,wang2025mindeditmllminsightdrivenediting} to explicit visual conditions (e.g., masks, edges)~\cite{controlnet_ref,mou2024t2i_refcontrol,ju2024brushnet} and interactive operations~\cite{mou2024diffeditor_drag,liu2024magicquill,shi2024dragdiffusion}. 
From the perspective of editing targets, these general-purpose methods excel at manipulating high-level semantic elements. They can be broadly categorized into three main streams: global image stylization, object-level manipulation, and local attribute modification. Global stylization methods~\cite{park2025stylestyletransfer,dai2025diffusefiststyletransfer} focus on altering the overall artistic style or illumination of an image while preserving the underlying structural layout. Object-level manipulation, which includes the insertion, removal, and replacement of entities, typically leverages masked inpainting techniques~\cite{rombach2021highresolutioninpainting} to seamlessly integrate new foreground objects into existing background contexts. Lastly, local attribute modification approaches aim to alter specific properties of an existing entity. More recently, the field has witnessed a paradigm shift towards general-purpose editing models~\cite{li2025uniworldedit,yu2025anyeditedit,lin2025uniworldedit,kulikov2025floweditedit,seedream2025seedreamedit,liu2025step1x-edit_textinstruction}, which can simultaneously and effectively execute these diverse manipulation tasks within a single unified framework.
\begin{figure*}[tb]
  \centering
  \includegraphics[width=0.95\textwidth]{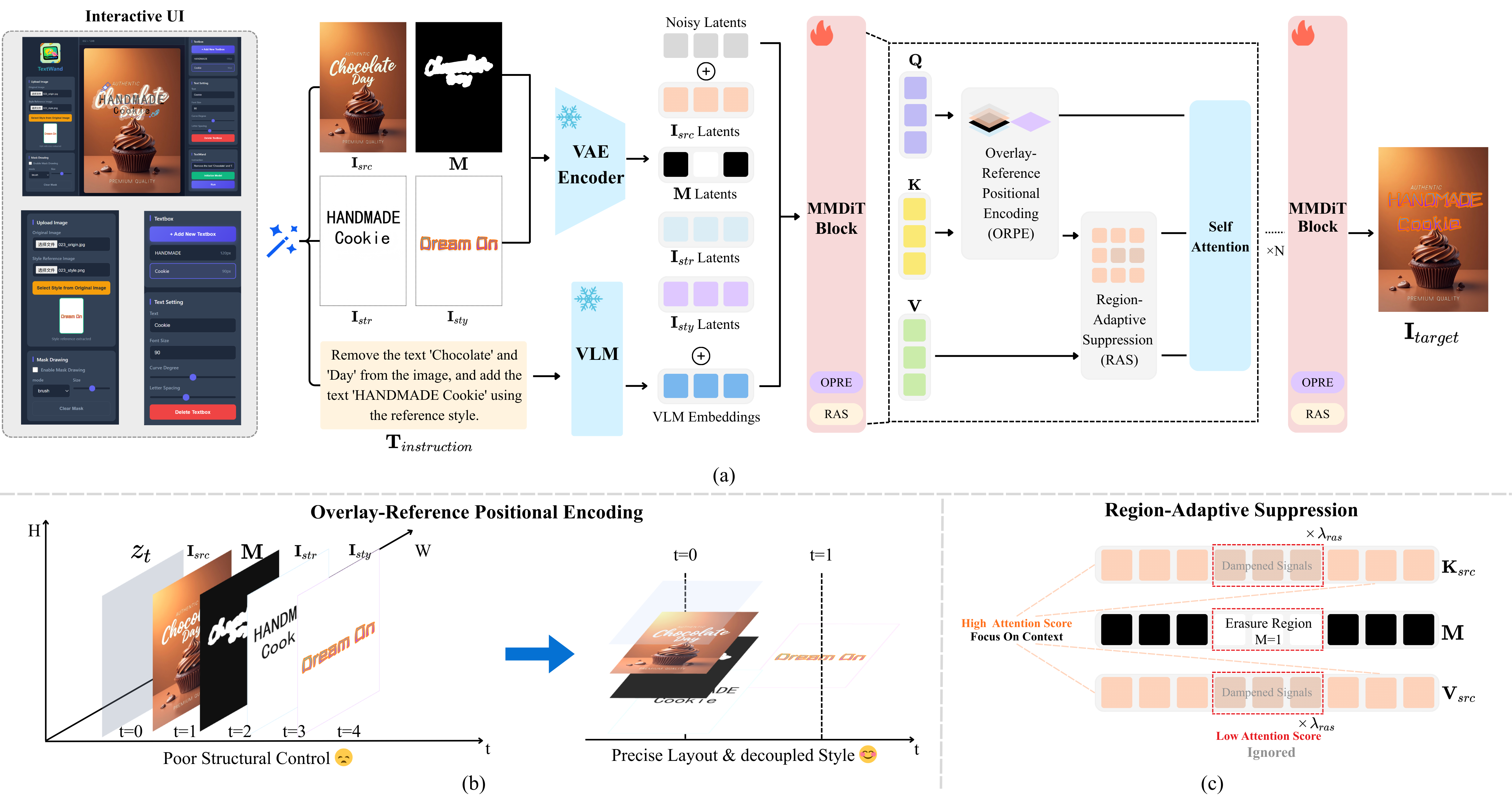}
\caption{The overview of our TextWand framework. 
    (\emph{a}) \textbf{Unified Editing Pipeline:} Through an interactive UI, the system takes a text instruction ($\mathbf{T}_{instruction}$) and specific visual conditions ($\mathbf{I}_{src}, \mathbf{M}, \mathbf{I}_{str}, \mathbf{I}_{sty}$) as inputs. This formulation explicitly decouples the editing operations: the mask $\mathbf{M}$ designates the region for text erasure, while the structure guide $\mathbf{I}_{str}$ controls the addition of new text. After encoding, stacked MMDiT blocks execute these decoupled erasure and rendering primitives within a single forward pass to output $\mathbf{I}_{target}$.
  (\emph{b}) Overlay-Reference Positional Encoding: Anchors the layout guide and source canvas at the same temporal coordinate for pixel-level alignment, while isolating the style reference.
  (\emph{c}) Region-Adaptive Suppression: Dampens the key/value signals ($\mathbf{K}_{src}, \mathbf{V}_{src}$) of the masked obsolete text, forcing self-attention to reconstruct a clean background from the unmasked context.
}
  \label{fig:overview}
\end{figure*}
\subsection{Scene Text Editing}
Scene text editing is harder than generic image editing because it must satisfy stricter constraints on both geometry and appearance: the text has to remain readable, follow precise typesetting layouts including perspective or curvature, and match coherent typography style, all while leaving the background intact.
Prior work~\cite{liu2024qtscenetextrecent,su2023scenescenetextrecent,ji2023improvingscenetextrecent,yang2024fontdiffuserscenetextrecent,nikolaidou2024diffusionpenscenetextrecent,8978022sceneedit,Fang_2025_CVPR,utdiff} has made notable progress, yet the current landscape remains fragmented. Recently, FonTS~\cite{fonts_iccv25} focuses on reference-driven typography styling, but it provides limited control over geometric typesetting and tends to be less reliable on complex real-world backgrounds. Content-centric methods such as FluxText~\cite{fluxtext_2505}, AnyText~\cite{tuo2023anytext}, AnyText2~\cite{anytext2_2411} and ControlText~\cite{controltext_2502} can generate or modify textual content, but they typically offer limited support for expressive typesetting, and their layouts are often unstable when long strings or challenging scenes are involved. EasyText~\cite{easytext_2505} improves geometric flexibility by supporting curved and arched layouts, yet it does not provide fine-grained control over typographic style to match a given reference. More unified attempts such as OmniText~\cite{omnitext_2510} move toward jointly handling multiple editing factors, but their controllability is still limited, for example to coarse attributes such as color, and they do not reliably support curved typesetting. Overall, existing methods either emphasize style at the expense of layout control, emphasize content and layout without reliable style transfer, or attempt unification with insufficient control granularity. This leaves an open need for a general-purpose scene text editing framework that can operate on complex backgrounds, support both deletion and insertion, and enable coordinated control of expressive layout and typographic style under simple user interactions.

\section{Methodology}
\label{sec:method}
\subsection{Overview of TextWand}
\label{sec:overview}

The goal of scene text editing is to manipulate text instances within an image while preserving background coherence. We characterize a text instance as $\mathcal{T}=\{C,L,S\}$, where $C$, $L$, and $S$ denote content, geometric layout, and visual style. We unify diverse editing tasks by reducing them to two atomic primitives: erasure $\mathcal{P}_{erase}$, which removes the textual evidence, and rendering $\mathcal{P}_{render}$, which synthesizes glyphs conditioned on layout and style controls.

TextWand adopts this decoupled formulation to handle scene text removal, generation and replacement within a single framework. We formulate the user intentions as explicit control signals. Let $\mathbf{I}_{src}$ denote the original source image to be edited, $\mathbf{M}$ denote the binary mask defining the region for erasure, $\mathbf{I}_{str}$ represents the structural skeleton, a geometrically neutral guide (e.g., standard font rasterization) that defines precise spatial positioning and glyph boundaries. Conversely, $\mathbf{I}_{sty}$ provides the exemplar style, containing the intricate typographic details, textures, and artistic effects. Using $\mathbf{I}_{str}$ and $\mathbf{I}_{sty}$ decouples \emph{where to place} from \emph{how to render}, enabling faithful style transfer onto a fixed geometric scaffold. Building upon these atomic primitives, the target image $\mathbf{I}_{target}$ produced by distinct editing tasks is formulated as specific combinations of these atomic primitives: for text removal, $\mathbf{I}_{target} = \mathcal{P}_{erase}(\mathbf{I}_{src}, \mathbf{M})$; for text generation, $\mathbf{I}_{target} = \mathcal{P}_{render}(\mathbf{I}_{src}, \mathbf{I}_{str}, \mathbf{I}_{sty})$; and for text replacement, $\mathbf{I}_{target} = \mathcal{P}_{erase}\circ \mathcal{P}_{render}(\mathbf{I}_{src}, \mathbf{I}_{str}, \mathbf{I}_{sty}, \mathbf{M})$, where $\circ$ represents the coordinated execution of erasure and rendering within the same inference pass, effectively steering the denoising process to satisfy both the structural constraints of $\mathcal{P}_{render}$ and the contextual consistency of $\mathcal{P}_{erase}$.

To realize these editing objectives while fundamentally avoiding the feature entanglement inherent in coupled approaches, we introduce \textbf{TextWand}. As illustrated in part (a) of~\ref{fig:overview}, we implement TextWand using a transformer-based diffusion architecture that integrates these primitives into a unified generative stream. Specifically, the visual inputs (source image~$\mathbf{I}_{src}$, mask~$\mathbf{M}$, structure~$\mathbf{I}_{str}$, and style reference~$\mathbf{I}_{sty}$) are first compressed into latents by a frozen VAE~\cite{kingma2013autovae}, while the text instruction~$\mathbf{T}_{instruction}$ is encoded into semantic embeddings via a frozen VLM. These condition latents and embeddings, alongside the noisy latents, are then fed into stacked Multi-Modal Diffusion Transformer (MMDiT) blocks. Within these blocks, rather than treating conditions as independent sequences, our method coordinates the editing process through two specialized mechanisms: Overlay-Reference Positional Encoding (ORPE) for precise spatial alignment with independent style referencing, and Region-Adaptive Suppression (RAS) to attenuate masked features during attention for clean erasure.
Furthermore, to ensure stable convergence across these diverse tasks, TextWand is optimized via a progressive curriculum training strategy. By decoupling the learning process, the model progressively advances from fundamental style transfer on clean backgrounds to complex text generation in real-world scenes, before finally integrating the erasure primitive for unified scene text editing. 
\subsection{Overlay-Reference Positional Encoding}
\label{sec:orpe}
We implement the proposed editing primitives by first encoding the input conditions into a sequence of latent feature tokens. Let $\mathbf{f}_{src}, \mathbf{f}_{str}, \mathbf{f}_{sty} \in \mathbb{R}^{L \times D}$ denote the token sequences extracted from the source image, structure image, and style reference, respectively, where $L$ is the sequence length and $D$ is the embedding dimension. In standard multi-modal diffusion transformers leveraging 3D Rotary Positional Embeddings (e.g., MS-RoPE~\cite{wu2025qwenimage}), multiple visual conditions are natively treated as independent, sequential instances along a temporal axis. As shown in part (b) of~\ref{fig:overview}(left), the model assigns a unique temporal index $t$ to each image input, formulating the position as $(t, h, w)$. Consequently, even if the structure image $\mathbf{f}_{str}$ and the source canvas $\mathbf{f}_{src}$ share the same spatial resolution $(H, W)$, they are projected into disjoint regions of the positional embedding space. This temporal separation burdens the attention mechanism to infer correspondences, leading to spatial drift in pixel-level layout adherence.

To overcome this limitation and enforce strict spatial correspondence for the rendering primitive, we introduce Overlay-Reference Positional Encoding (ORPE). Our core insight is to reconfigure the 3D coordinate system to introduce a task-specific inductive bias that reflects the logical alignment of inputs. We logically partition the inputs: the spatially-aligned canvas group ($f_{src}$, $f_{msk}$, $f_{str}$) and the geometrically independent reference group ($f_{sty}$). By forcibly overriding the sequential assignment and anchoring the canvas group to a shared 3D coordinate $p_{shared} = (t_0, h, w)$, ORPE injects a powerful geometric prior: the structural guide is perceived natively as a synchronized logical overlay, not a separate sequence. This explicit spatial synchronization guarantees coordinate-aligned self-attention, ensuring strict adherence to user-specified layouts. Conversely, isolating the style reference at $p_{style} = (t_{sty}, h, w)$ ($t_{sty} \neq t_0$) allows it to function as a floating semantic exemplar. 
This explicit spatial arrangement relieves the model from inferring complex long-range correspondences, ensuring that the generated text strictly adheres to the user-specified geometry while faithfully rendering the target style.

\subsection{Region-Adaptive Suppression}
\label{sec:ras}
The implementation of the erasure primitive $\mathcal{P}_{erase}$ faces a fundamental conflict, as the self-attention mechanism must gather background context from source image tokens while simultaneously ignoring the obsolete text features residing within the editing region. Without explicit intervention, standard attention layers naturally attend to these residual features, inevitably causing severe semantic artifacts in the final output.

We therefore propose Region-Adaptive Suppression (RAS) to explicitly shape the self-attention evidence field. Rather than using $\mathbf{M}$ as a passive binary hint, RAS selectively suppresses the $K_{src}$ and $V_{src}$ projected from $\mathbf{f}_{src}$ inside $\mathbf{M}$, while leaving non-masked context and other conditions fully accessible. This yields cleaner removal with minimal architectural change and is compatible with standard MMDiT blocks. Crucially, RAS is applied consistently across all denoising timesteps, maintaining a stable semantic vacuum that allows the rendering primitive to seamlessly composite new text features without temporal disruption. The detailed architectural flow of this operation is visualized in part (c) of~\ref{fig:overview}.

\textbf{Implementation.} 
Formally, let $K_{src}, V_{src} \in \mathbb{R}^{L \times D}$ denote the key and value matrices projected from the source image tokens. Before the softmax attention operation, we introduce a spatially-selective suppression mechanism:
\begin{equation}
    \hat{\textbf{K}}_{src}[p] = 
    \begin{cases} 
        \lambda_{ras} \cdot \textbf{K}_{src}[p] & \text{if } p \in \mathbf{M} \\ 
        \textbf{K}_{src}[p] & \text{otherwise} 
    \end{cases}
\end{equation}
\begin{equation}
    \hat{\textbf{V}}_{src}[p] = 
    \begin{cases} 
        \lambda_{ras} \cdot \textbf{V}_{src}[p] & \text{if } p \in \mathbf{M} \\ 
        \textbf{V}_{src}[p] & \text{otherwise} 
    \end{cases}
\end{equation}
where $p$ represents the spatial position index, and $\lambda_{ras} \in [0, 1)$ is a scalar suppression factor. 

This spatially-selective modulation reconfigures the attention manifold to eliminate persistent semantic interference. Consequently, the modified attention output is computed as:
\begin{equation}
    \text{Attention}(\textbf{Q}, \hat{\textbf{K}}, \hat{\textbf{V}}) = \text{softmax}\left(\frac{\textbf{Q}\hat{\textbf{K}}^\top}{\sqrt{d}}\right) \hat{\textbf{V}}
\end{equation}
By attenuating $K_{src}$, we effectively marginalize masked regions within the global affinity map, rendering obsolete features invisible to queries during the dot-product calculation. Simultaneously, the suppression of $V_{src}$ serves as a latent-level barrier to prevent residual feature leakage. This mechanism creates a semantic vacuum that forces the model to prioritize uncontaminated contextual priors and the precise structural guidance $\mathbf{f}_{str}$ for artifact-free reconstruction.

\subsection{Progressive Curriculum Training Strategy}
Simultaneous optimization of erasure and rendering often leads to severe feature entanglement between structural generation and background reconstruction. To address this, we propose a progressive curriculum that decomposes the learning process along a complexity gradient, allowing the model to acquire robust text rendering priors before tackling the complexities of unified contextual editing.

\textbf{Stage 1: Isolated Style Transfer.}
We first optimize the model to execute the rendering primitive ($\mathcal{P}_{render}$) using plain white backgrounds. This noise-free environment isolates style transfer via ORPE, ensuring precise typography and spatial typesetting without background interference.

\textbf{Stage 2: Context-Aware Typography Rendering.}
We then introduce real-world scenes but restrict the objective to the scene text generation task. Leveraging the typographic priors acquired in the first stage, the optimization shifts to environmental integration. The model learns to synthesize appropriate lighting, shadows, and textures to blend the generated text naturally into complex background contexts.

\textbf{Stage 3: Unified Task Integration.} 
Finally, we introduce the erasure primitive to jointly optimize the full spectrum of scene text editing tasks. Capitalizing on the robust rendering priors acquired in previous stages, the model effortlessly couples with RAS to seamlessly composite new text into cleanly erased backgrounds, achieving highly controllable editing.

\section{Experiments}
\label{sec:experiments}

\subsection{Experimental Settings}
\label{subsec:settings}

\begin{figure*}[t]
  \centering
  \includegraphics[width=0.95\textwidth]{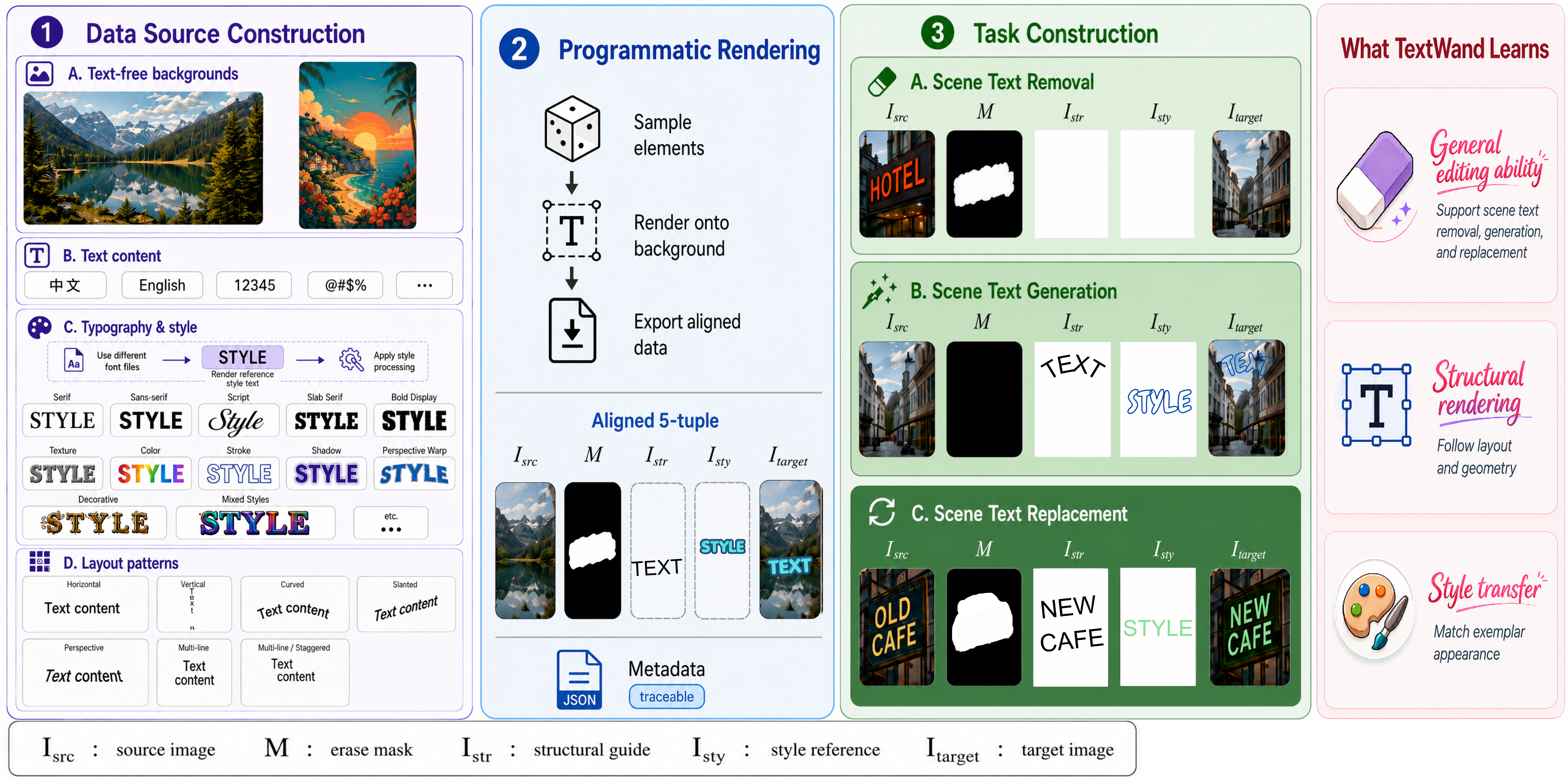}
  \caption{
  Overview of the TextWand data construction pipeline. 
  We first collect text-free backgrounds and diverse text/style/layout elements, then programmatically render aligned editing tuples, and finally construct task-specific samples for scene text removal, generation, and replacement.
  }
  \label{fig:data_construction_pipeline}
\end{figure*}

\textbf{Datasets and Benchmark.} 
A key obstacle to unified scene text editing lies in the lack of datasets that provide explicit and fine-grained supervision for controllable editing. Existing resources are often designed for isolated sub-tasks, such as text removal or text rendering, and typically provide only coarse text prompts or final edited images, without exposing the intermediate control signals required by professional editing workflows. As a result, they are insufficient for learning how to jointly control text content, spatial layout, typographic style, and background restoration. To bridge this fundamental infrastructure gap, we introduce TextWand-72K, a large-scale, unified dataset specifically architected for multi-attribute control. Built upon the Poster-100k dataset~\cite{chen2025postercraft}, it provides 72,000 high-quality training pairs featuring explicit, pixel-level structural guides and exemplar-driven style references. The dataset encompasses a diverse distribution of languages, layouts (including curved, slanted), and highly complex real-world background textures. To enable fair comparisons, we establish TextWand-Bench, consisting of 1,500 meticulously selected test samples evenly distributed across text removal, generation, and replacement (500 samples per task). To strictly prevent train-test leakage, all benchmark images are sourced from independent domains isolated from the Poster-100k training corpus, ensuring zero overlap. By providing standardized, high-precision control signals, TextWand-Bench offers a robust evaluation landscape for layout fidelity and style consistency. To facilitate future research and ensure full reproducibility, the TextWand-72K dataset, TextWand-Bench, and pre-trained model weights will be publicly released. For detailed construction process, please refer to the supplementary materials.
\begin{table*}[!t]
\centering
\caption{Quantitative results on the scene text generation and scene text replacement tasks. Arrows indicate whether higher ($\uparrow$) or lower ($\downarrow$) is better. The best results are highlighted in \textbf{bold}, and the second best are \underline{underlined}.}
\label{tab:quant_results_gen_rep}
\resizebox{0.9\linewidth}{!}{
\begin{tabular}{l ccccccc}
\toprule
\multirow{2}{*}{Method} & \multicolumn{7}{c}{Evaluation Metrics} \\
\cmidrule(lr){2-8}
& NED $\downarrow$ & $\text{IoU}_{\text{bbox}}$ $\uparrow$ & FID $\downarrow$ & LPIPS $\downarrow$ & SSIM $\uparrow$ & PSNR $\uparrow$ & CLIP-I $\uparrow$ \\
\midrule
\multicolumn{8}{c}{Scene Text Generation} \\
\midrule
AnyText2~\cite{anytext2_2411} & 0.9954 & 0.1852 & 87.15 & 0.2122 & 0.6722 & 17.06 & 0.8224 \\
FluxText~\cite{fluxtext_2505} & 0.3696 & \underline{0.7408} & \underline{49.43} & \underline{0.0579} & \underline{0.9165} & \underline{21.44} & \underline{0.9131} \\
Qwen-Image-Edit-2509~\cite{wu2025qwenimage} & 0.4782 & 0.6836 & 52.79 & 0.0955 & 0.8560 & 20.44 & 0.9017 \\
Seedream 4.5 & \underline{0.3361} & 0.2550 & 76.42 & 0.2624 & 0.6898 & 15.77 & 0.8677 \\
Nano Banana Pro & 0.4032 & 0.3436 & 52.02 & 0.1252 & 0.7585 & 20.28 & 0.9013 \\
TextWand (Ours) & \textbf{0.3319} & \textbf{0.8837} & \textbf{11.46} & \textbf{0.0325} & \textbf{0.9340} & \textbf{27.35} & \textbf{0.9654} \\
\midrule
\multicolumn{8}{c}{Scene Text Replacement} \\
\midrule
AnyText2~\cite{anytext2_2411} & 0.9938 & 0.1538 & 90.22 & 0.2179 & 0.6629 & 17.09 & 0.8204 \\
FluxText~\cite{fluxtext_2505} & 0.4629 & \underline{0.7115} & \underline{49.45} & \underline{0.0629} & \underline{0.9124} & \underline{21.36} & \underline{0.9056} \\
Qwen-Image-Edit-2509~\cite{wu2025qwenimage} & 0.4536 & 0.5821 & 51.84 & 0.0882 & 0.8988 & 20.38 & 0.8950 \\
Seedream 4.5 & \textbf{0.3199} & 0.5349 & 62.06 & 0.1770 & 0.7665 & 18.52 & 0.8816 \\
Nano Banana Pro & 0.3592 & 0.5341 & 52.32 & 0.1100 & 0.7623 & 20.54 & 0.9043 \\
TextWand (Ours) & \underline{0.3368} & \textbf{0.8721} & \textbf{14.03} & \textbf{0.0358} & \textbf{0.9291} & \textbf{26.87} & \textbf{0.9614} \\
\bottomrule
\end{tabular}
d}
\end{table*}

\textbf{Data Construction Pipeline.}
To obtain explicit and reproducible supervision for unified scene text editing, we design a programmatic data construction pipeline that synthesizes paired samples with fully controlled content, layout, style, mask, and target image. As shown in~\ref{fig:data_construction_pipeline}, 
each sample consists of five aligned components: the source image $\mathbf{I}_{src}$, the erasure mask $\mathbf{M}$, the structural guide $\mathbf{I}_{str}$, the style reference $\mathbf{I}_{sty}$, and the ground-truth target image $\mathbf{I}_{target}$, together with a metadata file recording all rendering parameters. 
This design allows us to train and evaluate the model under precise supervision rather than relying only on coarse natural language descriptions.

We start from text-free background images. 
For each sample, the text content is randomly sampled from a character pool containing over 5,000 Chinese characters, English letters, and common symbols. 
The typographic appearance is sampled from more than 300 font files and further diversified through randomized rendering effects. 
This produces a wide range of rasterized text styles, from simple flat typography to highly decorative poster-like text.

For layout construction, we generate both regular and expressive text arrangements. 
Horizontal text is randomly placed with varying font sizes, tracking, and small rotations. 
Curved text is generated by sampling cubic Bezier curves and placing characters according to arc length, which enables arched and slanted layouts. 
A trial-and-error validation procedure is used to reject invalid layouts whose glyph masks exceed image boundaries or violate placement constraints. 
This creates challenging replacement cases where the model must erase obsolete text and render new text in the same spatial region.

We construct three task types. 
For scene text generation, the source image is a clean background, the mask is empty, the structural guide is a black glyph mask rendered on a white canvas, the style reference is a centered exemplar rendered with the target style, and the ground truth is obtained by compositing the stylized target text onto the background. 
For scene text removal, we first render stylized text onto a clean background to form the source image, derive the erasure mask from the glyph region with dilation and blur, and use the original clean background as the ground truth. 
For scene text replacement, the source image contains the old stylized text, the mask is generated from the old text region, the structural guide specifies the new text layout, the style reference provides the desired appearance, and the ground truth is rendered by compositing the new stylized text onto the clean background. 
All masks are generated from exact glyph masks and then expanded to better cover strokes, shadows, and antialiasing artifacts, making the erasure supervision closer to practical user-provided brush masks.

The pipeline records all sampled parameters, including text strings, font identities, layout type, Bezier control points, character positions, rotation centers, bounding boxes, style attributes, and task instructions. 
Therefore, every sample is exactly traceable and can be regenerated or analyzed. 
By jointly varying text content, layout, and style between old and new text, our construction process provides diverse supervision for the three atomic editing abilities required by TextWand: clean erasure, faithful structural rendering, and exemplar-driven style transfer.

\textbf{Implementation Details.} 
Our method is built upon the Qwen-Image-Edit-2509~\cite{wu2025qwenimage} foundation model. We employ Low-Rank Adaptation (LoRA)~\cite{hu2022lora} to fine-tune the model, setting the LoRA rank to 128. The hyperparameter $\lambda_{ras}$ is set to 0.4. To ensure stable optimization, the process strictly follows the proposed progressive curriculum training strategy. Specifically, the model is trained for 1,000 steps in the first stage, 1,000 steps in the second stage, and 4,000 steps in the final stage, with a total training batch size of 64.

\begin{figure}[t]
\centering
  \includegraphics[width=\linewidth]{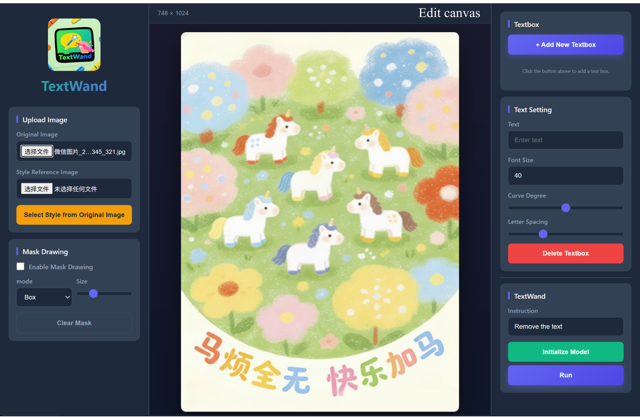}
  \caption{Screenshot of the interactive GUI for TextWand. The workspace allows users to effortlessly provide text prompts, draw masks, and guide the generation process within a unified pipeline.}
  \label{fig:ui_screenshot}
\end{figure}

\textbf{Interactive Graphical User Interface.}
\label{sec:ui_interface}To facilitate highly natural and intuitive scene text editing, we developed a comprehensive interactive Graphical User Interface (GUI) powered by our TextWand framework. As illustrated in Figure~\ref{fig:ui_screenshot}, the UI is designed to streamline the complex process of explicit control signal provision into a user-friendly workflow:
\begin{enumerate}
    \item \textbf{Input \& Erasure Module:} Users upload the source image $\mathbf{I}_{src}$ and can directly use a flexible brush tool to paint over obsolete text. This operation intuitively defines the binary erasure mask $\mathbf{M}$ for seamless text removal.
    \item \textbf{Layout \& Style Control Module:} To define the structure guide ($\mathbf{I}_{str}$), users type the target text and can freely manipulate its spatial layout into arbitrary forms, including horizontal, diagonal, arched, and freely rotated orientations. For the style reference ($\mathbf{I}_{sty}$), the UI offers dual modes: users can either upload an external exemplar or conveniently draw a bounding box to extract in-context typography directly from the source image. If no reference is provided, the system seamlessly defaults to contextual blending.
    \item \textbf{Execution Module:} With a single click, the unified framework coordinates the erasure and rendering primitives ($\mathcal{P}_{erase} \diamond \mathcal{P}_{render}$) within the same inference pass to generate the final highly realistic target image.
\end{enumerate}

This interactive system demonstrates the practical applicability, high degree of freedom, and robust zero-shot generalization of TextWand in real-world graphic design and photo editing scenarios.

\begin{figure*}[tb]
  \centering
  \includegraphics[width=0.95\textwidth]{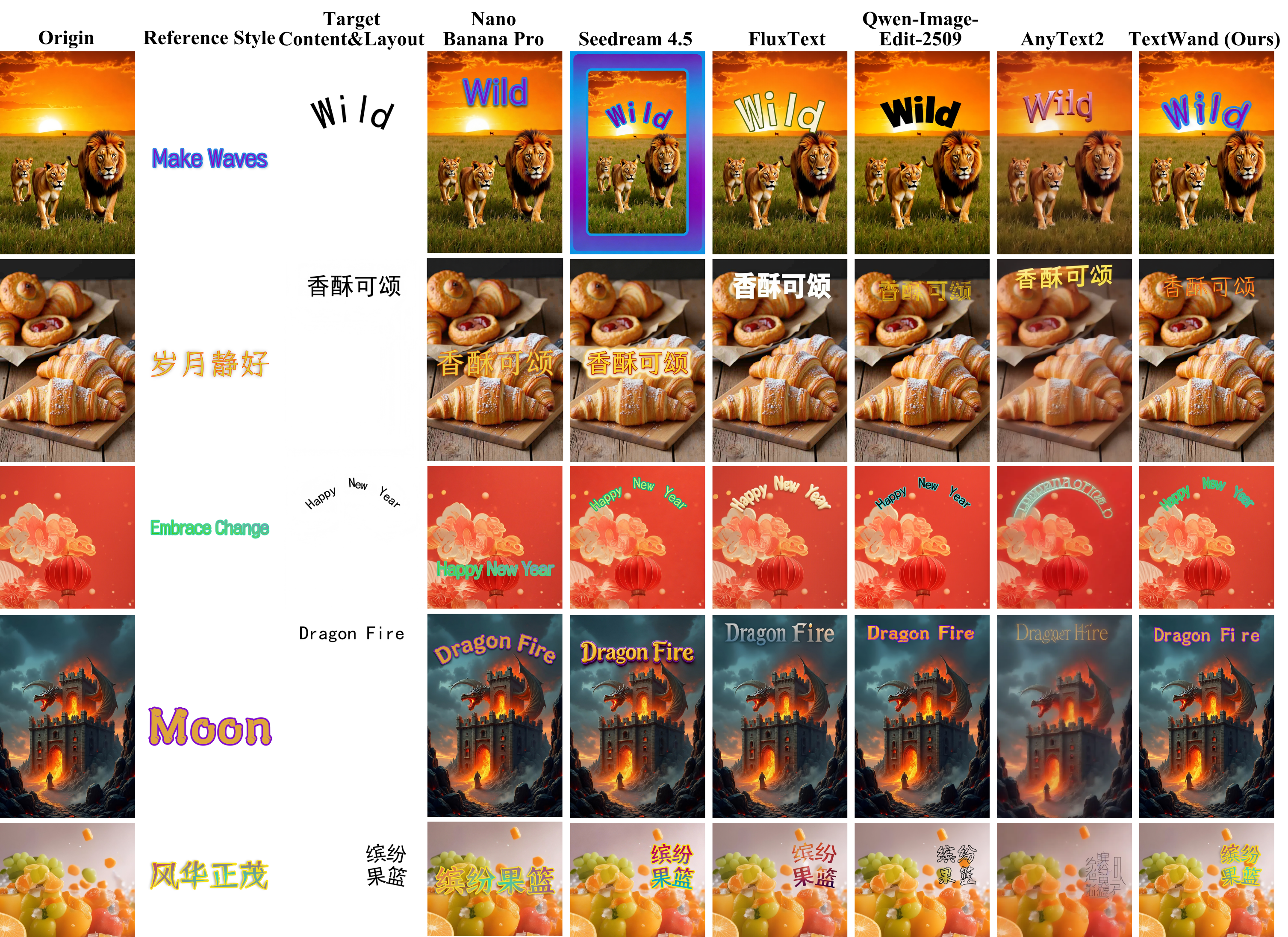}
  \caption{Qualitative comparison on the scene text generation task. 
  }
  \label{fig:qualitative_generation}
\end{figure*}

\subsection{Qualitative and Quantitative Results}
\label{subsec:results}
\begin{figure*}[!t]
  \centering
  \begin{subfigure}{0.3\linewidth}
    \includegraphics[width=\linewidth]{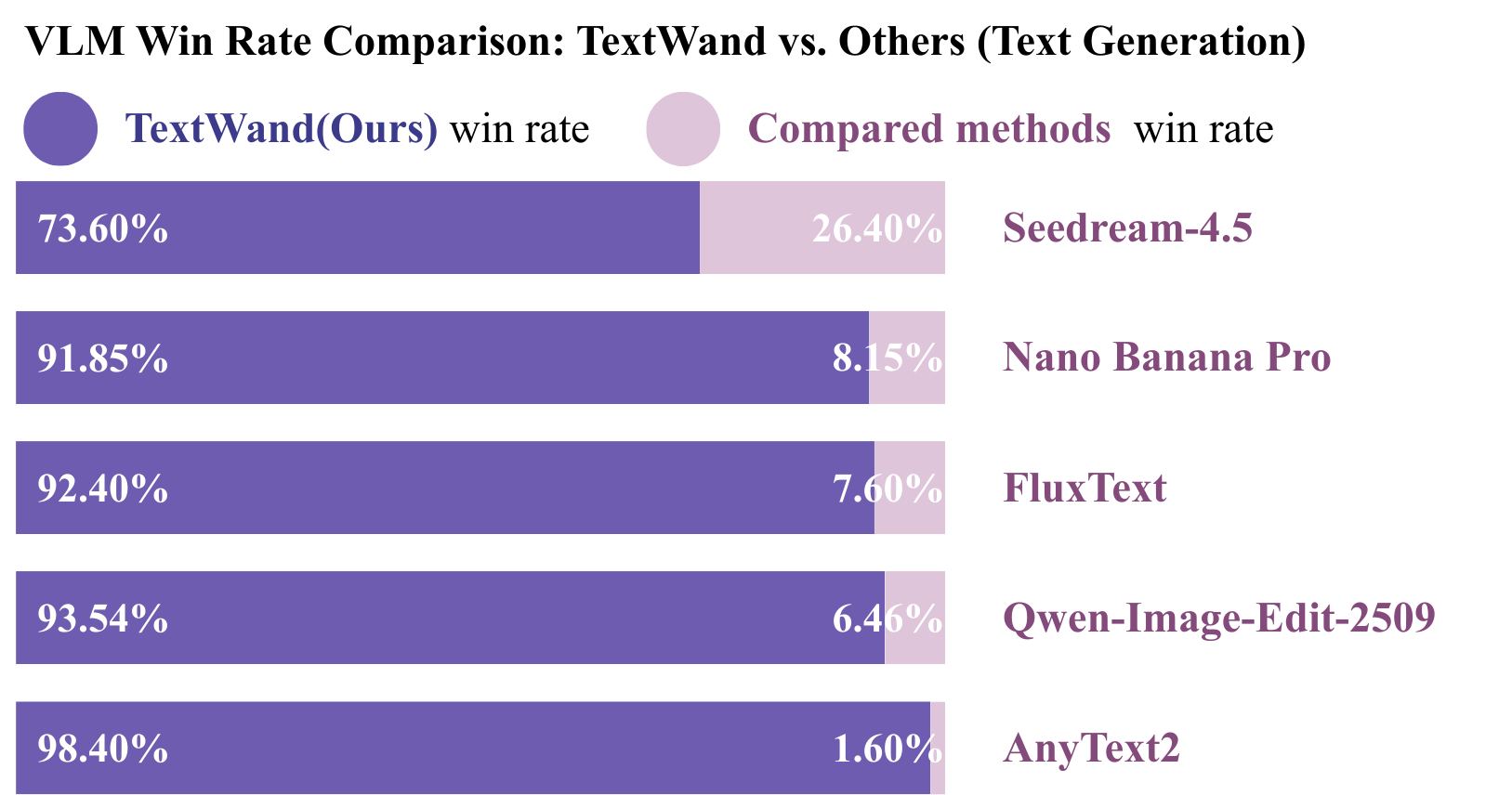}
    \caption{Text Generation}
    \label{fig:winrate_addition}
  \end{subfigure}
  \hfill
  \begin{subfigure}{0.3\linewidth}
    \includegraphics[width=\linewidth]{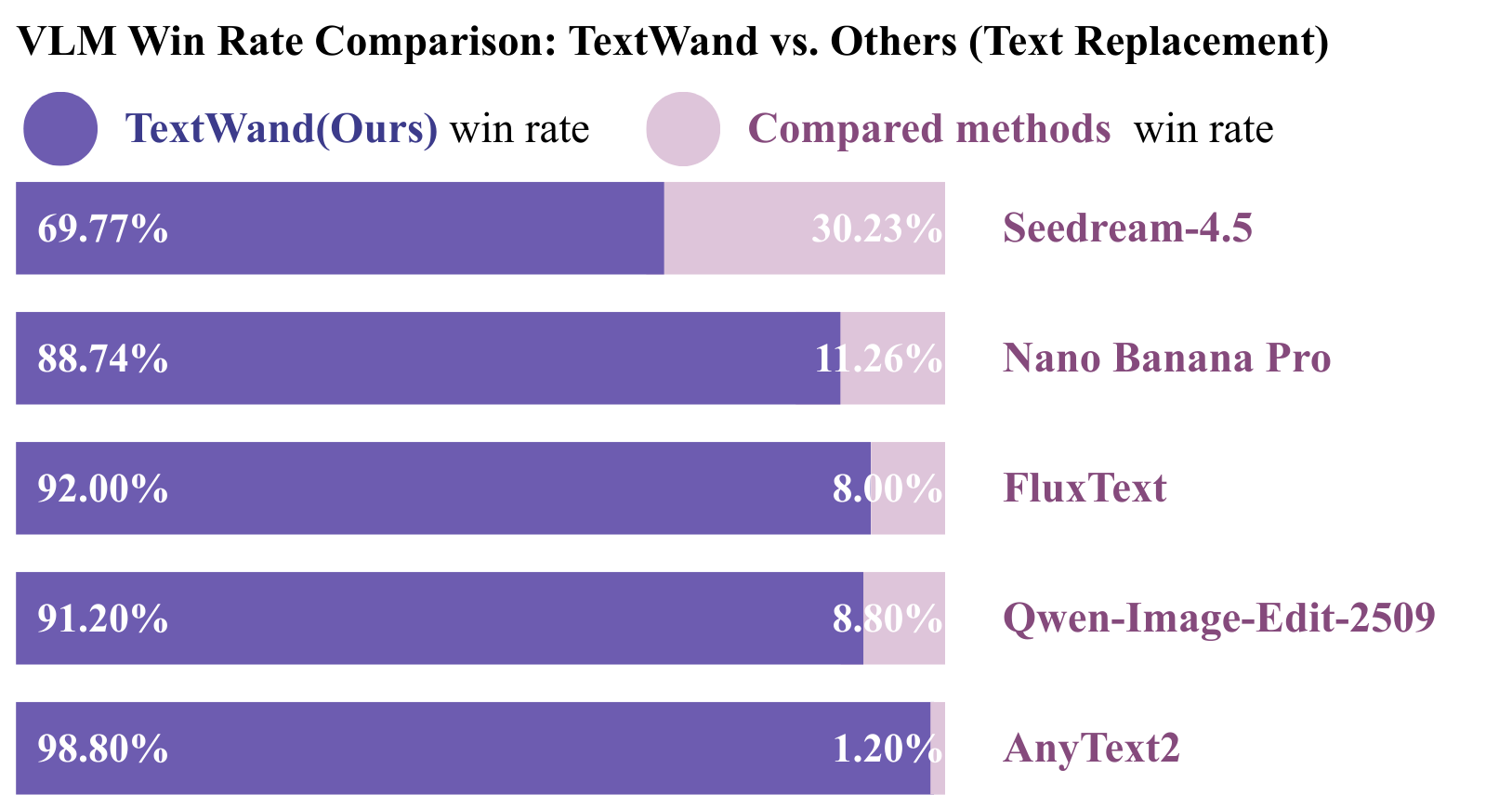}
    \caption{Text Replacement}
    \label{fig:winrate_replacement}
  \end{subfigure}
  \hfill
  \begin{subfigure}{0.34\linewidth}
    \includegraphics[width=\linewidth]{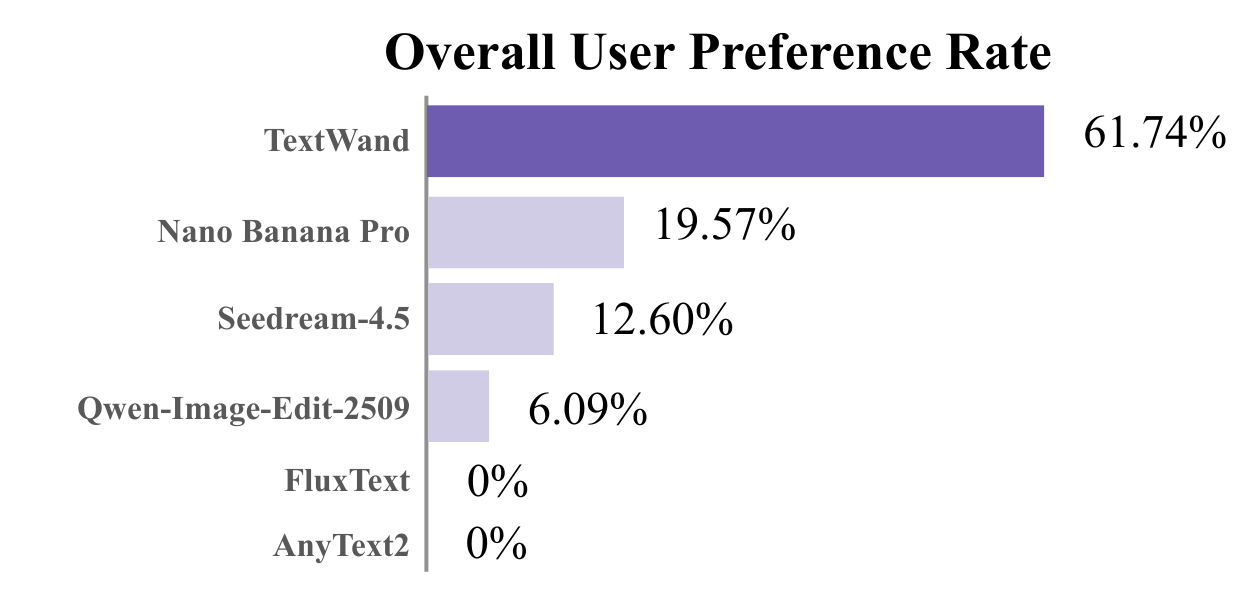}
    \caption{User Study}
    \label{fig:user_study_subplot}
  \end{subfigure}
  \caption{Visual style assessment and human preference. VLM win rates on the (a) scene text generation and (b) scene text replacement tasks. (c) Overall preference rate from the user study.}
  \label{fig:winrate_comparison}
\end{figure*}

\textbf{Evaluation Metrics}. To align with the stringent requirements of scene text editing discussed earlier, we evaluate the editing performance across three distinct sets of metrics.
(1) Image Quality and Background Consistency: We utilize FID , LPIPS~\cite{zhang2018unreasonablelpips}, SSIM~\cite{wang2004imagessim}, PSNR and CLIP-I~\cite{radford2021learningintro} to assess the perceptual quality and visual fidelity. These metrics are applied across all tasks, serving as the primary indicators.
(2) Text Content and Layout Accuracy: For text generation and replacement tasks, we adopt Normalized Edit Distance (NED) to evaluate the OCR-level text accuracy, and Bounding Box IoU ($\text{IoU}_{\text{bbox}}$) to measure the precision of spatial layout control.
(3) Style Consistency: Traditional metrics often struggle to capture fine-grained typographic styles. Following recent advancements in Large Vision-Language Models (VLMs) as visual evaluators, we employ Qwen3-VL-8B~\cite{bai2025qwen3vl} as an objective judge. Given a triplet consisting of the ground truth (GT), our generated result, and the result of a comparison method, the VLM is instructed to select the image that better matches the font style, color, and texture of the GT. We randomize pair ordering and blind method identities when querying the judge to reduce evaluation bias. In~\ref{fig:winrate_comparison}, we report the win rate of our method to quantify style transfer performance.

\begin{figure*}[tb]
  \centering
  \includegraphics[width=0.98\textwidth]{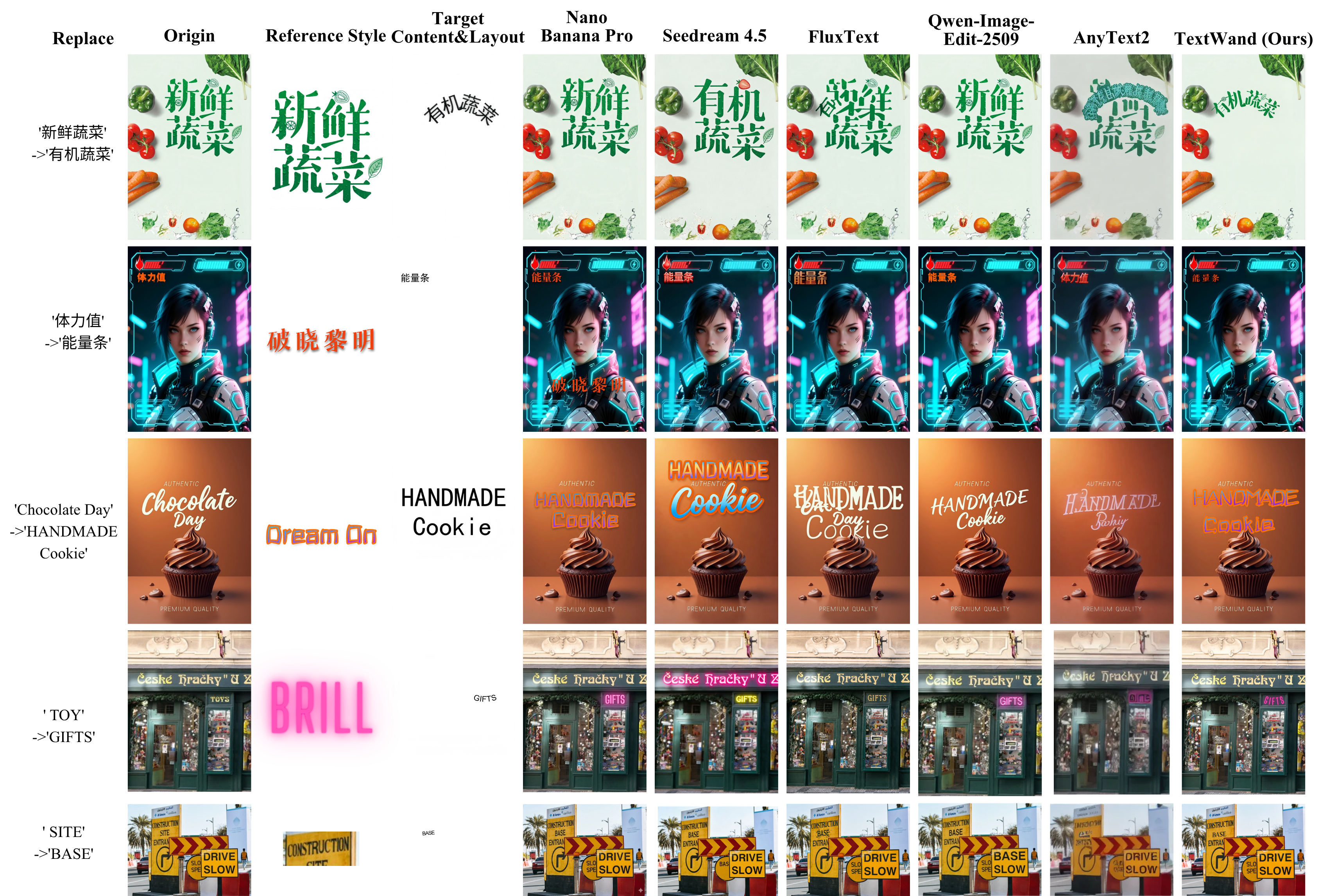}
  \caption{Qualitative comparison on the scene text replacement task.}
  \label{fig:qualitative_replacement}
\end{figure*}

\textbf{Comparative Methods.}
Due to the absence of existing unified frameworks for comprehensive scene text editing, we construct task-specific comparison groups. Across all three tasks (removal, generation, and replacement), we benchmark against leading general-purpose image editing models, encompassing both open-source (Qwen-Image-Edit-2509~\cite{wu2025qwenimage}, FLUX.1-Kontext~\cite{labs2025flux1kontextflowmatching}, LongCat~\cite{LongCat-Image}) and closed-source (Nano Banana Pro, Seedream 4.5) systems. For the rendering-intensive generation and replacement tasks, we additionally incorporate the latest text-centric models, AnyText2~\cite{anytext2_2411} and FluxText~\cite{fluxtext_2505}. To maintain a rigorous evaluation protocol, our chosen baselines represent the current state-of-the-art in handling explicit spatial guidance or localized text rendering. Methods lacking architectural support for fine-grained region editing fundamentally fail at these precise tasks, yielding trivial comparisons. Thus, they are excluded to ensure our benchmark remains strictly focused on competitive models.

\textbf{Scene Text Generation.} 
Scene text generation demands simultaneous adherence to strict content, layout, and style constraints. As detailed in~\ref{tab:quant_results_gen_rep}, TextWand establishes new state-of-the-art performance across all perceptual image quality metrics and secures a commanding win rate in the VQA-based style assessment (\ref{fig:winrate_addition}). These quantitative achievements are visually evident in~\ref{fig:qualitative_generation}. While all comparative methods systematically suffer from severe spatial misalignments and incorrect typographic styles, TextWand flawlessly executes the user intent in every instance. In contrast to these competing approaches, our method strictly anchors to the provided layout guidance and accurately renders the target text with the exact desired lighting, color, and font style, consistently delivering highly realistic in-the-wild editing without exception.

\begin{table}[t]
\centering
\caption{Quantitative results on the scene text removal task.}
\label{tab:quant_results_removal}
\resizebox{0.98\linewidth}{!}{
\begin{tabular}{l ccccc}
\toprule
\multirow{2}{*}{Method} & \multicolumn{5}{c}{Evaluation Metrics} \\
\cmidrule(lr){2-6}
& FID $\downarrow$ & LPIPS $\downarrow$ & SSIM $\uparrow$ & PSNR $\uparrow$ & CLIP-I $\uparrow$ \\
\midrule
FLUX.1-Kontext~\cite{labs2025flux1kontextflowmatching} & 60.16 & 0.0852 & \underline{0.9106} & 23.59 & 0.8726 \\
LongCat-Image-Edit~\cite{LongCat-Image} & 25.85 & 0.0853 & 0.7418 & 22.34 & 0.9440 \\
Qwen-Image-Edit-2509~\cite{wu2025qwenimage} & \underline{21.32} & 0.0686 & 0.8100 & 25.35 & 0.9538 \\
Seedream 4.5 & 31.80 & 0.1475 & 0.7724 & 23.94 & 0.9415 \\
Nano Banana Pro & \textbf{19.53} & \underline{0.0549} & 0.7940 & \underline{25.48} & \textbf{0.9586} \\
TextWand (Ours) & 25.84 & \textbf{0.0498} & \textbf{0.9155} & \textbf{27.28} & \underline{0.9577} \\
\bottomrule
\end{tabular}
}
\end{table}
\begin{figure*}[!tb]
  \centering
  \includegraphics[width=0.98\linewidth]{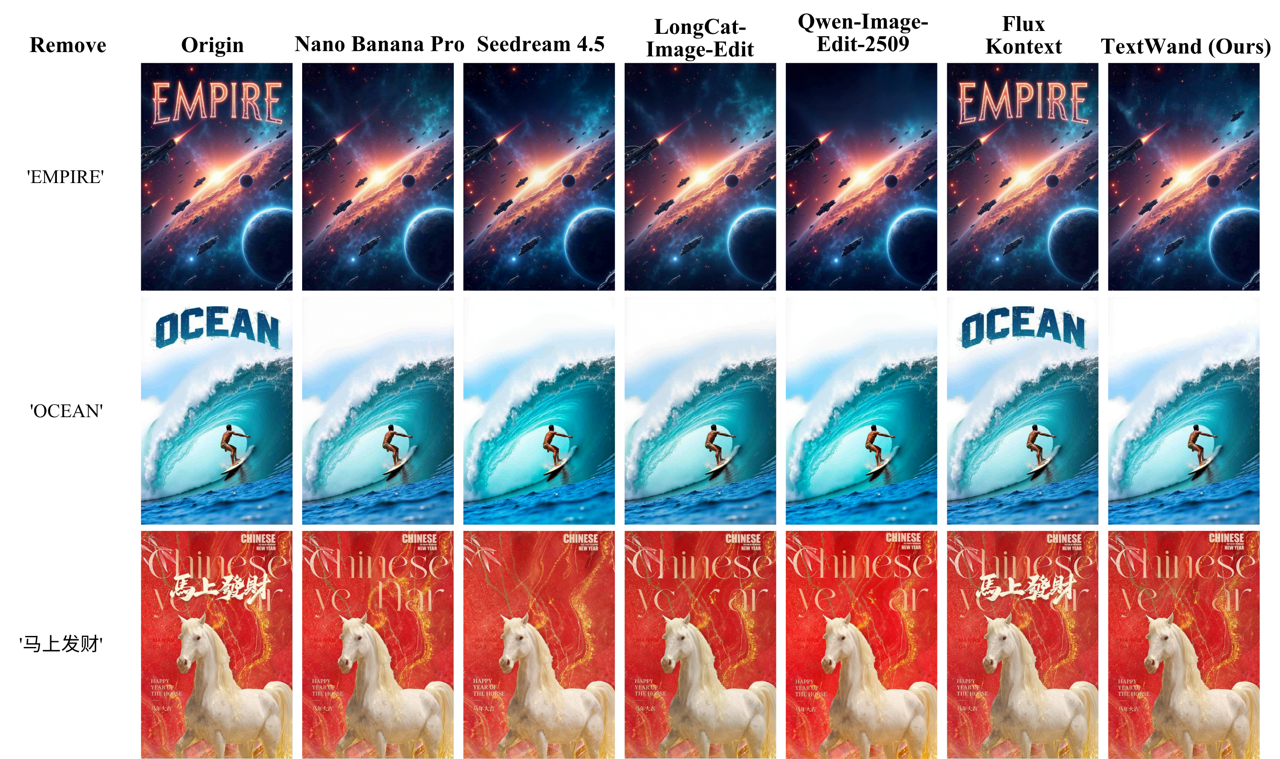}
  \caption{Qualitative comparison on the scene text removal task. 
  }
  \label{fig:qualitative_removal}
\end{figure*}
\textbf{Scene Text Replacement.} 
Scene text replacement presents the formidable dual challenge of entirely erasing legacy text while synthesizing new content under strict layout and style constraints. As detailed in~\ref{tab:quant_results_gen_rep}, TextWand consistently outperforms all baselines across perceptual image quality metrics, demonstrating superior background preservation. Furthermore, our method achieves a commanding win rate in the VQA-based style assessment (\ref{fig:winrate_replacement}). As shown in~\ref{fig:qualitative_replacement}, our method achieved the best visual results compared to other methods. By fundamentally decoupling erasure from rendering, TextWand successfully eliminates the residual ghosting artifacts that severely limit prior methods, achieving unprecedented stylistic fidelity.

\textbf{Scene Text Removal.} 
Scene text removal requires eliminating outdated texts while perfectly reconstructing occluded backgrounds. As demonstrated in~\ref{tab:quant_results_removal}, our method outperforms other comparative methods across most of image quality metrics, and maintains a marginal performance gap in the remaining few. This quantitative advantage is further corroborated by our qualitative results in~\ref{fig:qualitative_removal}. FLUX.1-Kontext~\cite{labs2025flux1kontextflowmatching} method fails to remove the target text. Nano Banana Pro method generates additional anomalous characters, while the Seedream 4.5 method excessively erases adjacent non-target text. In contrast, our approach seamlessly reconstructs complex background textures without affecting the surrounding visual context.

\textbf{User Study.}
A blind 6-AFC user study (23 participants, 10 diverse cases) demonstrates TextWand's commanding 61.74\% preference rate (\ref{fig:user_study_subplot}). This overwhelming human consensus validates our visual superiority and highly realistic syntheses, definitively offsetting the aforementioned OCR metric discrepancies. Questionnaire details are in the \textit{supplementary materials}.

\begin{table*}[t]
\centering
\caption{Quantitative ablation study of TextWand on the scene text generation and replacement tasks.}
\label{tab:ablation}
\resizebox{0.8\linewidth}{!}{
\begin{tabular}{l ccccccc}
\toprule
\multirow{2}{*}{Method} & \multicolumn{7}{c}{Evaluation Metrics} \\
\cmidrule(lr){2-8}
& NED $\downarrow$ & $\text{IoU}_{\text{bbox}}$ $\uparrow$ & FID $\downarrow$ & LPIPS $\downarrow$ & SSIM $\uparrow$ & PSNR $\uparrow$ & CLIP-I $\uparrow$ \\
\midrule
\multicolumn{8}{c}{Scene Text Generation} \\
\midrule
w/o ORPE & 0.3777 & \underline{0.8710} & \underline{12.24} & \underline{0.0330} & \textbf{0.9340} & \underline{27.23} & \underline{0.9635} \\
w/o RAS & \underline{0.3769} & 0.8651 & 12.90 & 0.0341 & \underline{0.9332} & 26.95 & 0.9630 \\
Train in Single Stage & 0.4152 & 0.8503 & 18.67 & 0.0402 & 0.9284 & 25.66 & 0.9516 \\
TextWand (Ours) & \textbf{0.3319} & \textbf{0.8837} & \textbf{11.46} & \textbf{0.0325} & \textbf{0.9340} & \textbf{27.35} & \textbf{0.9654} \\
\midrule
\multicolumn{8}{c}{Scene Text Replacement} \\
\midrule
w/o ORPE & \underline{0.3653} & \underline{0.8718} & \textbf{13.92} & \underline{0.0372} & \underline{0.9253} & \underline{26.73} & \textbf{0.9615} \\
w/o RAS & 0.3986 & 0.8665 & 15.30 & 0.0384 & 0.9245 & 26.46 & 0.9590 \\
Train in Single Stage & 0.4415 & 0.8223 & 22.08 & 0.0451 & 0.9246 & 24.90 & 0.9445 \\
TextWand (Ours) & \textbf{0.3368} & \textbf{0.8721} & \underline{14.03} & \textbf{0.0358} & \textbf{0.9291} & \textbf{26.87} & \underline{0.9614} \\
\bottomrule
\end{tabular}
}
\end{table*}
\begin{figure*}[!tb]
  \centering
  \includegraphics[width=0.8\linewidth]{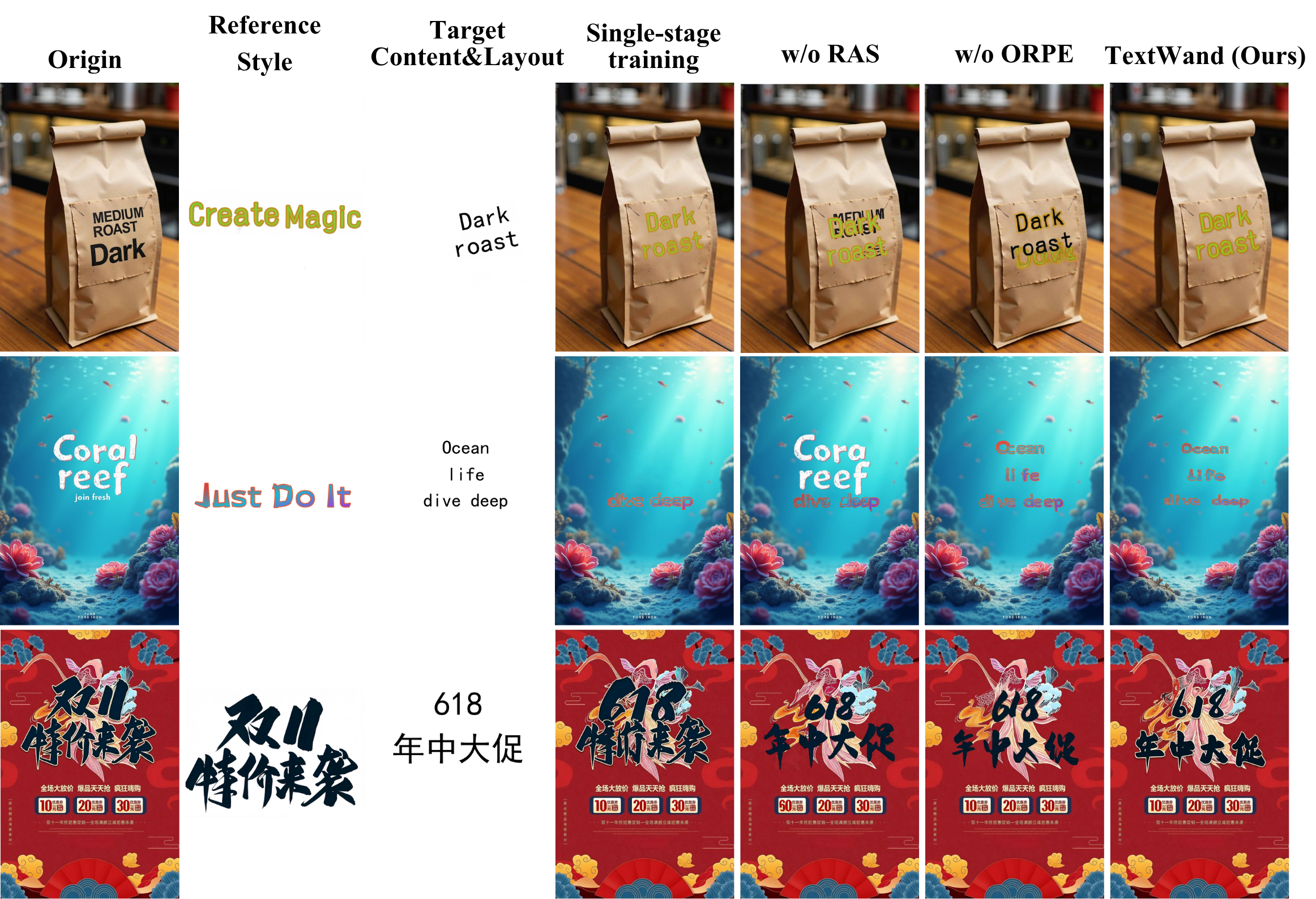}
  \caption{Qualitative ablation study of TextWand.}
  \label{fig:qualitative_ablation}
\end{figure*}

\subsection{Ablation Study}
\label{subsec:ablation}

To validate the necessity of our proposed components, we conduct extensive ablation studies on the text generation and replacement tasks, evaluating the model under three degraded settings. The quantitative results are summarized in~\ref{tab:ablation}. The qualitative results are displayed in~\ref{tab:ablation}.

To evaluate the effectiveness of the proposed ORPE, we replace it with the original sequential positional encoding, MSRoPE, in our base model Qwen-Image-Edit-2509. As shown in~\ref{tab:ablation}, removing ORPE in the scene text generation and replacement tasks severs the explicit spatial correspondence between the structure guidance and the canvas. Forced to rely on implicit long-range dependencies, the model suffers a substantial decline across the majority of evaluation metrics. Furthermore, qualitative results in~\ref{fig:qualitative_ablation} visually corroborate this finding, demonstrating that the model without ORPE frequently generates text with incorrect scales and misaligned positions, utterly failing to adhere to the provided structural constraints. This widespread performance drop confirms that ORPE is crucial for enforcing joint layout-style control.

To evaluate the effectiveness of the proposed RAS, we remove this suppression mechanism from our model. As reported in~\ref{tab:ablation}, without RAS, the self-attention mechanism fails to suppress the original text and inevitably aggregates these obsolete features. This feature entanglement leads to a drastic degradation in image quality metrics. Furthermore, qualitative comparisons in~\ref{fig:qualitative_ablation} visually corroborate this decline, revealing that the variant without RAS produces severe overlapping artifacts in the edited regions. These widespread quantitative and qualitative degradations firmly highlight the indispensable role of RAS in decoupling outdated semantic signals for clean, artifact-free editing.
    
To evaluate the effectiveness of the proposed progressive curriculum training strategy, we train the model directly from scratch without this stage-wise strategy. As reported in~\ref{tab:ablation}, abandoning the curriculum leads to severe optimization conflicts between structural generation and texture blending. This instability causes a significant performance drop across all quantitative metrics. Furthermore, qualitative comparisons in~\ref{fig:qualitative_ablation} visually corroborate this decline, revealing that the model trained with a single-stage strategy exhibits weaker preservation of the reference text style and often suffers from missing or incomplete characters, indicating that directly learning layout construction and style rendering together leads to unstable optimization. The variant without RAS fails to fully erase the original text, leaving visible remnants that interfere with the final editing result. Meanwhile, removing ORPE causes the generated text t lose the desired reference style; instead, its appearance tends to remain close to the original text. This variant also produces more frequent spatial misalignment, with the rendered text deviating from the target layout. These qualitative degradations demonstrate that progressive curriculum training, robust text erasure, and explicit style-layout disentanglement are all essential for achieving accurate and visually consistent scene text editing.

\section{Conclusion}
\label{sec:conclusion}
We propose TextWand, a unified scene text editing framework that introduces ORPE for precise layout-style control and RAS for seamless background reconstruction. Empowered by our custom dataset and progressive curriculum, it achieves state-of-the-art performance across text removal, generation, and replacement without exhaustive engineering optimizations, highlighting its immense scaling potential.
\section{Limitations}
\label{sec:limitation}
Although TextWand demonstrates strong performance in unified scene text editing, several limitations remain. 
First, the current framework has not been fully optimized for inference efficiency. Future work can investigate model distillation, quantization, efficient sampling, and lightweight attention designs to accelerate inference.
Second, deploying TextWand on resource-constrained devices remains challenging. 
For mobile-side applications such as on-device poster editing, e-commerce image retouching, and interactive photo editing, it is important to further reduce model size while preserving text legibility, layout fidelity, and reference-style consistency. 
Adapting the proposed ORPE and RAS mechanisms to smaller generative backbones is therefore a promising direction.

\bibliography{sn-bibliography}

\end{document}